\def\editmode{0}

\def\bibfilenames{bibman_refs,refs}

\def\spsformat{1}

\if\spsformat1

\documentclass{article}
\usepackage{spconf}
\usepackage{balance}

\else
\documentclass[11pt,final,onecolumn]{IEEEtran}

\fi

\usepackage{savesym} 
\savesymbol{labelindent}
\usepackage{enumitem} 

\if\editmode1  
\usepackage[backend=bibtex,style=alphabetic,sorting=debug,maxbibnames=99]{biblatex}
\DeclareFieldFormat{labelalpha}{\thefield{entrykey}}
\DeclareFieldFormat{extraalpha}{}
\bibliography{\bibfilenames}
\newcommand{\cmt}[1]{\noindent\textcolor{lightgreen}{\underline{[#1]}}} 
\newcommand{\hc}[1]{\textcolor{blue}{#1}} 

\setlistdepth{9}
\newlist{bulletlist}{enumerate}{9}
\setlist[bulletlist,1]{label=$\bullet$}
\setlist[bulletlist,2]{label=$\diamond$}
\setlist[bulletlist,3]{label=$\rightarrow$}
\setlist[bulletlist,4]{label=$\circ$}
\setlist[bulletlist,5]{label=$-$}
\setlist[bulletlist,6]{label=$\square$}
\setlist[bulletlist,7]{label=$\star$}
\setlist[bulletlist,8]{label=$\checkmark$}
\setlist[bulletlist,9]{label=$\Delta$}
\newenvironment{bullets}{\begin{bulletlist}}{\end{bulletlist}}

\newcommand{\blt}[1][noargpassed]{
  \item%
  \ifthenelse{\equal{#1}{noargpassed}}{}{\cmt{#1}}%
}

\else
\usepackage{cite}
\newcommand{\cmt}[1]{} 
\newcommand{\hc}[1]{\textcolor{black}{#1}} 
\newenvironment{bullets}{}{}
\newcommand{\blt}[1][noargpassed]{\ignorespaces}

\fi

\newcommand{\printmybibliography}{
\if\editmode1 
\printbibliography
\else
\bibliography{\bibfilenames}
\fi
}

\usepackage{fixltx2e}

\usepackage{graphicx}

\usepackage{subcaption}              

\usepackage[utf8]{inputenc}

\usepackage{amsfonts}
\usepackage{amsmath}
\usepackage{mathtools}

\usepackage{amssymb}

\usepackage{bm}

\usepackage{color,verbatim}
\usepackage{multirow}
\usepackage{accents}

\usepackage{theoremref}

\usepackage{url}

\newcounter{rulecounter}
\newcommand{\resetrule}{ \setcounter{rulecounter}{0}}
\resetrule

\newtheorem{myauxproblem}{Problem}

\newtheorem{myauxoptionalproblem}{Optional Problem}

\newsavebox{\selvestebox}
\newenvironment{colbox}[1]
  {\newcommand\colboxcolor{#1}%
   \begin{lrbox}{\selvestebox}%
   \begin{minipage}{\dimexpr\columnwidth-2\fboxsep\relax}}
  {\end{minipage}\end{lrbox}%
   \begin{center}
   \colorbox{\colboxcolor}{\usebox{\selvestebox}}
   \end{center}}

\definecolor{orange}{rgb}{1,0.8,0}
\definecolor{gray}{rgb}{.9,0.9,0.9}
\definecolor{darkgray}{rgb}{.3,0.3,0.3}
\definecolor{darkblue}{rgb}{.1,0.0,0.3}
\definecolor{lightblue}{rgb}{0.7,0.7,1}
\definecolor{lightred}{rgb}{1,0.7,.7}
\definecolor{purple}{RGB}{204,153,255}
\definecolor{lightgray}{rgb}{.95,0.95,0.95}
\definecolor{lightgreen}{rgb}{0.3,0.5,0.3}
\definecolor{darkgreen}{rgb}{0.05,0.3,0.05}

\newcommand{\hbm}[1]{{\hat{\bm #1}}}

\newcommand{\rfield}{\mathbb{R}}

\newcommand{\transpose}{^\top}
 \newcommand{\define}{:=}

\newcommand{\prob}{\mathop{\textrm{P}} }

\newcommand{\expected}{\mathop{\textrm{E}}\nolimits}

\newcommand{\hz}{{\mathcal{H}_0}}
\newcommand{\ho}{{\mathcal{H}_1}}

\newtheorem{myproposition}{Proposition}
\newtheorem{myremark}{Remark}
\newtheorem{myproblemstatement}{Problem Statement}
\newtheorem{mylemma}{Lemma}
\newtheorem{mytheorem}{Theorem}
\newtheorem{mydefinition}{Definition}
\newtheorem{mycorollary}{Corollary}

\renewcommand{\expected}{\hc{\mathbb{E}}}
\renewcommand{\prob}{\hc{\mathbb{P}}}

\newcommand{\Region}{\hc{\mathcal{X}}}
\newcommand{\Loc}{\hc{\bm x}} 

\newcommand{\TxSig}{\hc{s}} 
\newcommand{\RxSig}{\hc{r}} 
\newcommand{\Cha}{\hc{h}} 
\newcommand{\Noi}{\hc{z}} 

\newcommand{\Rss}{\hc{f}}
\newcommand{\EstRss}{\hc{\hat f}}
\newcommand{\VecRss}{\hc{\bm f}}
\newcommand{\EstVecRss}{\hc{\hbm f}}

\newcommand{\NotFea}[1]{\hc{_{#1}}} 
\newcommand{\IndFea}{\hc{n}} 

\newcommand{\NumFea}{\hc{N}} 
\newcommand{\IndSam}{\hc{k}} 
\newcommand{\NumSam}{\hc{K}} 
\newcommand{\SamPer}{\hc{T}} 

\newcommand{\IndDat}{{\hc{m}}} 
\newcommand{\NumDat}{\hc{M}} 

\newcommand{\Tra}{\hc{_\text{tr}}} 
\newcommand{\Val}{\hc{_\text{val}}} 

\newcommand{\IndPai}{{\hc{p}}} 
\newcommand{\NumPai}{\hc{P}} 

\newcommand{\IndEst}{{\hc{i}}} 
\newcommand{\NumEst}{\hc{I}} 
\newcommand{\NotEst}[1]{^{(#1)}} %

\newcommand{\DatSam}{\hc{\mathcal{D}_\text{s}}} 
\newcommand{\DatDif}{\hc{\mathcal{D}_\text{d}}} 

\newcommand{\Dec}{\hc{d}} 
\newcommand{\Net}{\hc{G}} 
\newcommand{\AuxNet}{\hc{\tilde G}} %
\newcommand{\AuxNetWei}{\hc{\tilde G}_{\VecWei}} %
\newcommand{\VecWei}{\hc{\bm w}} 
\newcommand{\NetWei}{\hc{G}_{\VecWei}} 
\newcommand{\Lik}{\hc{\mathcal{L}}} 

\newcommand{\Thr}{\hc{\gamma}} 
\newcommand{\NumClu}{\hc{C}} 

\newcommand{\cbr}[1]{\textcolor{black}{#1}} 
\usepackage{hyperref}

\begin{document}

\title{Spoofing Attack Detection in the Physical Layer\\ with
Commutative Neural Networks}

\if\spsformat1
\name{Daniel Romero$^1$, Peter Gerstoft$^2$, Hadi Givehchian$^2$, Dinesh Bharadia$^2$\thanks{
This research has been funded in part by the Research Council of Norway under IKTPLUSS grant 311994 and Intelligence Advanced Research Projects Activity (IARPA), via 2021-2106240007.}
\thanks{Data and code to reproduce
the experiments are available at \url{https://github.com/fachu000/spoofing}}}
\address{$^1$Dept. of ICT, University of Agder, Norway. Email: daniel.romero@uia.no \\
$^2$Electrical and Computer Engineering, University of California, San Diego, USA  }
\else
\author{Author(s) Name(s)\thanks{Thanks to XYZ agency for funding.}}
\fi

\maketitle

\begin{abstract}
In a spoofing attack, an attacker impersonates a legitimate user to
access or tamper with data intended for or produced by the legitimate
user. In wireless communication systems, these attacks may be detected
by relying on features of the channel and transmitter radios. In this
context, a popular approach is to exploit the dependence of the
received signal strength (RSS) at multiple receivers or access points
with respect to the spatial location of the transmitter. Existing
schemes rely on long-term estimates, which makes it difficult to
distinguish spoofing  from movement of a legitimate user.
This limitation is here addressed by means of a deep neural network
that implicitly learns the distribution of pairs of short-term RSS
vector estimates. The adopted network architecture imposes the
invariance to permutations of the input (commutativity) that the
decision problem exhibits. The merits of the proposed algorithm are
corroborated on a data set that we collected.


\end{abstract}

\begin{keywords}
Spoofing attack, wireless networks, deep learning.
\end{keywords}

\section{Introduction}
\label{sec:intro}

\begin{bullets}
\blt[Motivation]
\begin{bullets}
\blt[General]The pervasive presence of wireless communications in
virtually all human activities has spurred  the proliferation of
techniques that seek to illicitly obtain private data, compromise the
uptime of remote services, or impersonate other
users~\cite{kolias2015intrusion,vanhoef2016mac,martin2019handoff,tippenhauer2011requirements}.
\blt[spoofing]Spoofing attacks, which pursue the latter goal, are
especially problematic since they endow the attacker with the capacity
to access and modify data intended for or produced by a legitimate
user.
\blt[detecting spoofing] Detecting this kind of attacks is therefore
of paramount importance to guarantee data security. 

\blt[PHY]To hinder the ability of an attacker to impersonate a
legitimate user, cryptographic techniques are employed in various
communication layers, from the medium access control (MAC) layer to
the application layer. However, these security barriers are readily
bypassed by an attacker with the user credentials. For this reason,
many techniques to detect spoofing attacks in the physical layer have
been developed, typically by exploiting physical layer characteristics
that are specific to the transmitter, such as carrier frequency offset
(CFO), or features of the channel, such as received signal strength
(RSS) or Angle of Arrival (AoA). \cbr{These spoofing detection
techniques generally match observed features to previously collected
ones for a given legitimate transmitter.}

\end{bullets}

\blt[Literature]
\begin{bullets}
\blt[imperfections]More specifically, imperfections such as CFO, I/Q
offset, and I/Q imbalance, which uniquely characterize the analog
hardware of transmitters, have been used to verify user identity
in~\cite{brik2008wireless,givehchian2022evaluating,liu2019real,vo2016fingerprinting}. Unfortunately,
these techniques need knowledge of the communication protocol of the
transmitter and may fail under changes in the environment; e.g. in 
temperature~\cite{givehchian2022evaluating}.
\blt[AoA, TDoA, SNR]These limitations have been partially alleviated in
\cite{xiong2010secureangle,xiong2013securearray,shi2016robust}, which
employ AoA and TDoA features, and in~\cite{wang2020machine}, where a
neural network is trained on signal-to-noise ratio (SNR) traces
obtained in the sector level sweep process in mm-Wave 60 GHz IEEE
802.11ad Networks. However, they still require synchronization and/or
knowledge of the communication protocol.

\blt[RSS]In comparison, RSS-based techniques do not need to know
protocols or decode signals, which greatly enhances their generality
and applicability for spoofing
detection~\cite{chen2007detecting,yang2012detection,xiao2016phy,zeng2011identity,alotaibi2016new}. The
most common technique in this context is to apply K-means to RSS
measurements collected by multiple
receivers~\cite{chen2007detecting,hoang2018soft,sobehy2020csi}.  The
idea is to leverage the dependence of RSS signatures on the location
of the transmitter (and potential attacker) to declare an attack iff
transmissions with the same user ID (as provided by higher
communication layers) are determined to be originated at different
locations. Unfortunately, this means that attacks are declared
\emph{even when a legitimate user transmits from different locations
because it moves}. Thus, to satisfactorily tell spoofing from motion,
one needs to employ a \emph{higher-level algorithm} that utilizes the
decisions of the aforementioned \emph{low-level} RSS-based detectors
for multiple pairs of transmissions. To see the intuition, suppose for
example that one transmission is received from point A, then B, then
C, and then D, where all these spatial locations are declared to be
different by the \emph{low-level} algorithm. This suggests that the
user is moving. In turn, if the low-level algorithm determines that
A$=$C and B$=$D, then it is most likely that the legitimate user is at
A and an attacker at B or vice versa.
\blt[limitations]Unfortunately, such (low-level) algorithms
necessitate accurate RSS measurements, which in turn require long
averaging time intervals.  This limits their time resolution and,
therefore, their usage by higher-level algorithms to distinguish
spoofing from movement. 
%
\end{bullets}

\blt[contribution]This observation calls for RSS-based approaches
capable of effectively solving the aforementioned low-level decision
problem with RSS estimates that average just a small number of
received samples. To understand the challenge, recall that RSS is
generally estimated by averaging the squared magnitude of samples of
the received signal. For a large number of samples, these estimates
converge to the actual RSS. Unlike existing works, which generally
assume converged RSS estimates, a method with high temporal resolution
must rely on short-term averages, which randomly fluctuate around the
true RSS. Developing such a method is the main contribution of the
present paper. To this end, the distribution of such noisy RSS
estimates is implicitly learned in a data-driven fashion using a deep
neural network (DNN). The adopted architecture abides by the
commutative nature of the decision problem. The data collection
process just involves recording a small number of samples
corresponding to a few tens of different transmitter locations. The
high accuracy of the proposed algorithm is evaluated on a data set of
real measurements that we collected as part of this research work.

\end{bullets}

\section{Model and Problem Formulation}
\label{sec:mpf}

\cmt{model}
\begin{bullets}
\blt[Space]Let $\Region\subset \rfield^3$ index the spatial region of
interest, where all transmitters, both legitimate users and
attackers, are located.
\blt[RSS as a function of position]A transmitter at a fixed position
$\Loc\in\Region$
\begin{bullets}
\blt[tx. signal] transmits a signal whose complex baseband equivalent
version is $\TxSig(t)$, where $t$ denotes time.  This signal,
modeled as an unknown wide-sense stationary stochastic process,
\blt[rx. signal]is received by $\NumFea$ static receivers, which may
be, e.g., access points or base stations. After downconversion, the
signal at the 
$\IndFea$-th receiver reads as
\begin{align}
        \RxSig\NotFea{\IndFea}(\Loc,t) = \Cha\NotFea{\IndFea}(\Loc,t)\ast \TxSig(t) + \Noi\NotFea{\IndFea}(t),
\end{align}
where $\ast$ denotes convolution, $\Cha\NotFea{\IndFea}(\Loc,t)$ is
the unknown deterministic impulse response of the
bandpass equivalent channel  between location $\Loc$ and the
$\IndFea$-th receiver, and $\Noi\NotFea{\IndFea}(t)$ is noise.
\blt[RSS]As usual, $\Noi\NotFea{\IndFea}(t)$ is assumed wide-sense stationary and
uncorrelated with $\TxSig(t)$, which implies that the mean-square
magnitude of $\RxSig\NotFea{\IndFea}(\Loc,t)$ is independent on $t$. Thus,
one can define the RSS of signal plus noise as
\begin{align}
        \Rss\NotFea{\IndFea}(\Loc) \define 10\log_{10} \expected{|
\RxSig\NotFea{\IndFea}(\Loc,t)
|^2},
\end{align}
where $\expected$ denotes expectation. 
\end{bullets}%
\blt[RSS estimates]To estimate $\Rss\NotFea{\IndFea}(\Loc)$, the
$\IndFea$-th receiver averages the square magnitude of $\NumSam$ samples of $\RxSig\NotFea{\IndFea}(\Loc,t)$, that is:
\begin{align}
\label{eq:estrss}
\EstRss\NotFea{\IndFea}(\Loc)\define 10\log_{10}
\sum_{\IndSam=0}^{\NumSam-1}
|\RxSig\NotFea{\IndFea}(\Loc,\IndSam \SamPer)|^2,
\end{align}
where $\SamPer$ is the sampling interval. If
$\RxSig\NotFea{\IndFea}(\Loc,\IndSam \SamPer)$ is ergodic for each $\Loc$, as
typically assumed, it follows that $\EstRss\NotFea{\IndFea}(\Loc)$
converges to $\Rss\NotFea{\IndFea}(\Loc)$ when $\NumSam\rightarrow
\infty$.

\blt[RSS vector]A fusion center or central controller forms the RSS
vector estimate $\EstVecRss(\Loc)\define
[\EstRss\NotFea{0}(\Loc),\ldots,\EstRss\NotFea{\NumFea-1}(\Loc)]\transpose$
by gathering these $\NumFea$ RSS estimates. It is assumed that the
position $\Loc$ of the transmitter does not change significantly
during the process of acquiring and collecting these estimates, which
is a mild assumption if $\NumSam$ is small, as considered here.
\end{bullets}

\cmt{problem formulation}With this notation, one can readily formulate
the problem as follows.
\begin{bullets}%
\blt[Given]Suppose that an RSS vector estimate is obtained for two
transmissions and let $\Loc_1,\Loc_2\in \Region$ denote the
(potentially equal) locations where the transmissions have been
originated.  Given $\EstVecRss(\Loc_1)$ and $\EstVecRss(\Loc_2)$, the
problem is to
\blt[Decide]decide between the following hypotheses:
\begin{align}
\label{eq:test}
\begin{cases}
\hz:~\Loc_1=\Loc_2
\\
\ho:~\Loc_1\neq \Loc_2.
\end{cases}
\end{align}
\blt[Dataset]To assist in this task, a data set of feature vectors
$\{\EstVecRss\NotEst{\IndEst}(\Loc_\IndDat),$ ${\IndDat=0\ldots
\NumDat-1},~\IndEst=0,\ldots,\NumEst-1\}$ is given, where
$\Loc_\IndDat\neq \Loc_{\IndDat'}$ for all $\IndDat \neq \IndDat'$ and
$\{\EstVecRss\NotEst{\IndEst}(\Loc_\IndDat)\}_{\IndEst=0}^{\NumEst-1}$ denote $\NumEst$ estimates of 
$\VecRss(\Loc_\IndDat)\define [
\Rss\NotFea{0}(\Loc_\IndDat),\ldots,\Rss\NotFea{\NumFea-1}(\Loc_\IndDat)]\transpose$.
\end{bullets}

\section{DNN-based Spoofing Detector}

\cmt{overview}This section describes the architecture and training
process of the proposed detector. 

\subsection{Data Set}
\label{sec:data}

\begin{bullets}
\blt[Form pairs] As indicated in Sec.~\ref{sec:mpf}, the given data
set comprises vectors of the form
$\EstVecRss\NotEst{\IndEst}(\Loc_\IndDat)$. To train a DNN-based
detector, a data set of pairs of such vectors needs to be
constructed with both pairs that correspond to the same transmitter
location and with pairs that correspond to different transmitter
locations. If one wishes to minimize the probability of error, as
described later, it is desirable that the resulting data set contains
the same number $\NumPai$ of pairs from each of these categories. The
procedure is formally described next. 
\begin{bullets}

\blt[Same points]First, for $\IndPai=0,\ldots,\NumPai-1$, 
\begin{bullets}
\blt draw $\IndDat_\IndPai$ uniformly at random from $\{0,\ldots,\NumDat-1\}$ and draw
\blt$\IndEst_\IndPai$ and 
$\IndEst'_\IndPai$ uniformly at random without replacement (i.e. no
replacement \emph{for each} $\IndPai$) from $\{0,\ldots,\NumEst-1\}$.
\end{bullets}
Then, form
\begin{align*}
\DatSam   \define \{&
(\EstVecRss\NotEst{\IndEst_\IndPai}(\Loc_{\IndDat_\IndPai}),
\EstVecRss\NotEst{\IndEst'_\IndPai}(\Loc_{\IndDat_\IndPai})),\\&~\IndPai=0,\ldots,\NumPai-1
\}\subset\rfield^{\NumFea} \times \rfield^{\NumFea}.
\end{align*}

\blt[Different points]Similarly, for $\IndPai=0,\ldots,\NumPai-1$, 
\begin{bullets}
\blt[] draw $\IndDat_\IndPai$ and  $\IndDat'_\IndPai$ uniformly at random without replacement (again for each $\IndPai$) from $\{0,\ldots,\NumDat-1\}$ and draw
\blt[]$\IndEst_\IndPai$ and 
$\IndEst'_\IndPai$ uniformly at random without replacement (i.e. no
replacement \emph{for each} $\IndPai$) from $\{0,\ldots,\NumEst-1\}$.
\end{bullets}
Then, form
\begin{align*}
\DatDif   \define \{&
(\EstVecRss\NotEst{\IndEst_\IndPai}(\Loc_{\IndDat_\IndPai}),
\EstVecRss\NotEst{\IndEst'_\IndPai}(\Loc_{\IndDat'_\IndPai})),\\&~\IndPai=0,\ldots,\NumPai-1
\}\subset\rfield^{\NumFea} \times \rfield^{\NumFea}.
\end{align*}
\end{bullets}
\end{bullets}

\subsection{Architecture}
\begin{bullets}
\blt[Overview]

\blt[Test] A  detector (or binary classifier) is a function
$\Dec:\rfield^{\NumFea} \times \rfield^{\NumFea} \rightarrow
\{\hz,\ho\}$ that returns a hypothesis for each input
$(\EstVecRss,\EstVecRss')$.
\blt[Test statistic]In the signal processing terminology, such a
detector is constructed by setting $\Dec(\EstVecRss,\EstVecRss')=\ho$
iff $\Net(\EstVecRss,\EstVecRss')$ exceeds a certain threshold, where
$\Net:\rfield^{\NumFea} \times \rfield^{\NumFea} \rightarrow \rfield$
is termed \emph{detection statistic}. The probability of error is minimized when
\begin{align}
\label{eq:logits}
\Net(\EstVecRss,\EstVecRss') = \log\frac{
\prob[\ho|\EstVecRss,\EstVecRss']
}{
\prob[\hz|\EstVecRss,\EstVecRss']
}=\log\frac{
\prob[\ho|\EstVecRss,\EstVecRss']
}{
1-\prob[\ho|\EstVecRss,\EstVecRss']
}, 
\end{align}
and the threshold is 0~\cite[Sec.~3.7]{kay2}. Satisfying
\eqref{eq:logits} is not generally possible using a finite data set,
but a function $\Net(\EstVecRss,\EstVecRss')$ can be learned to
approximately satisfy \eqref{eq:logits}, as described in
Sec.~\ref{sec:training}. The threshold to be used with the learned
function will still be 0.

\blt[DNN]
\begin{bullets}
\blt[Commutativization] Observe that, if one implements $\Net$
directly as a DNN $\NetWei$, where $\VecWei$ is the vector of
parameters such as weights and offsets, it will generally happen that
$\NetWei(\EstVecRss,\EstVecRss')\neq\NetWei(\EstVecRss',\EstVecRss)$,
which is inconsistent with the symmetry of \eqref{eq:test}. Thus, the
proposed approach is instead to use a DNN to implement an auxiliary
function $\AuxNetWei$ and then set
\begin{align}
\label{eq:commuta}
\NetWei(\EstVecRss,\EstVecRss') = \frac{
\AuxNetWei(\EstVecRss,\EstVecRss') + \AuxNetWei(\EstVecRss',\EstVecRss)
}{2}. 
\end{align}
It can be easily seen that
$\NetWei(\EstVecRss,\EstVecRss')=\NetWei(\EstVecRss',\EstVecRss)$ and
that the 2 in the denominator \eqref{eq:commuta} can be absorbed into
$\AuxNetWei(\EstVecRss,\EstVecRss')$.

\blt[Each component]Function $\AuxNetWei$ will be implemented as a
composition of more fundamental functions called layers. The following
is a description of the layers used in our experiments, but other
possibilities can be considered.
\begin{bullets}
\blt[First layer]The first layer $ \AuxNet_1(\EstVecRss,\EstVecRss')$
is non-trainable and complements the input with $\NumFea$ linear
features to facilitate learning:
\begin{align}
   \AuxNet_1(\EstVecRss,\EstVecRss') = 
 [\EstVecRss,\EstVecRss']   \left[
\begin{array}{c c c}
1&0&1\\
0&1&-1
\end{array}
\right].
\end{align}
\blt[Hidden layers]The subsequent 3 (hidden) layers are fully
connected with leaky ReLU activations and 512 neurons~\cite{goodfellow2016deep}. 
\blt[Output layer]Finally, the output layer is also fully connected
and has a single neuron with a linear activation.

\end{bullets}

\end{bullets}

\end{bullets}

\subsection{Training}
\label{sec:training}

\begin{bullets}
\blt[sigmoid]From \eqref{eq:logits}, it follows that
\begin{align}
\prob[\ho|\EstVecRss,\EstVecRss']=
\frac{1}
{1 + \exp[{-\NetWei(\EstVecRss,\EstVecRss')}]}
\define \sigma( \NetWei(\EstVecRss,\EstVecRss')),
\end{align}
where $\sigma$ denotes the so-called sigmoid function.
\blt[likelihood]Given the way the data set was constructed, it follows
that the log-likelihood of $\VecWei$ is given by
\begin{align}
\begin{aligned}
\label{eq:likelihood}
  \Lik(\VecWei;\DatSam,\DatDif)=&\sum_{(\EstVecRss,\EstVecRss')\in
  \DatSam}\log(1-\sigma( \NetWei(\EstVecRss,\EstVecRss'))
  \\&+\sum_{(\EstVecRss,\EstVecRss')\in
  \DatDif}\log(\sigma( \NetWei(\EstVecRss,\EstVecRss')).
  \end{aligned}
\end{align}

\blt[SGD]Following standard practice, a local minimum of the loss
function $ -\Lik(\VecWei;\DatSam,\DatDif)$ can be efficiently
approximated using stochastic gradient
descent~\cite{goodfellow2016deep}. 

\blt[Data]To avoid overfitting, a common technique is to apply an
early-stopping approach, where optimization is halted when the
\emph{validation accuracy} stops increasing. To this end, the given
$\NumDat$ transmitter locations are split into $\NumDat\Tra$ locations
for training and $\NumDat\Val\define \NumDat-\NumDat\Tra$ locations for validation. From the
first, $2\NumPai\Tra$ pairs are constructed as explained in
Sec.~\ref{sec:data} and substituted in \eqref{eq:likelihood} to obtain
the training loss. On the other hand, $2\NumPai\Val$ pairs are
obtained from the $\NumDat\Val$ validation locations to obtain the
validation accuracy, i.e., the number of validation pairs successfully
classified.

\blt[sparsity]It was observed that an $\ell_1$ regularizer on the
weights of the first trainable layer improves learning, likely because
the variability across space of some features is too chaotic and,
thus, not informative. Hence, promoting sparsity in this layer
encourages the DNN to select the  most informative features.

\end{bullets}

\section{Experiments}

\cmt{Overview} This section empirically validates the proposed
algorithm using real data. A link to the code and
data set is provided on the first page.
\cmt{simulation setup}%
\begin{bullets}%
\blt[data collection]%
\begin{bullets}%
\blt[room] Data collection took place in a room of the University of
California, San Diego, as depicted in Fig.~\ref{fig:room}.
\blt[transmitter] A transmitter was sequentially placed at 52
locations. For the sake of reproducibility, the transmitted signal
$\TxSig(t)$ is a 5 MHz sinusoid at a carrier frequency of 2.3 GHz.
\blt[receivers]For every transmitter, four receivers with four
antennas each record 4888 samples of the received signal. Thus,
 $\NumFea$ can be set between 1 and~4$\cdot$4=16. 

\end{bullets}

\begin{figure}[t]
 \centering
 \includegraphics[width=0.45\textwidth]{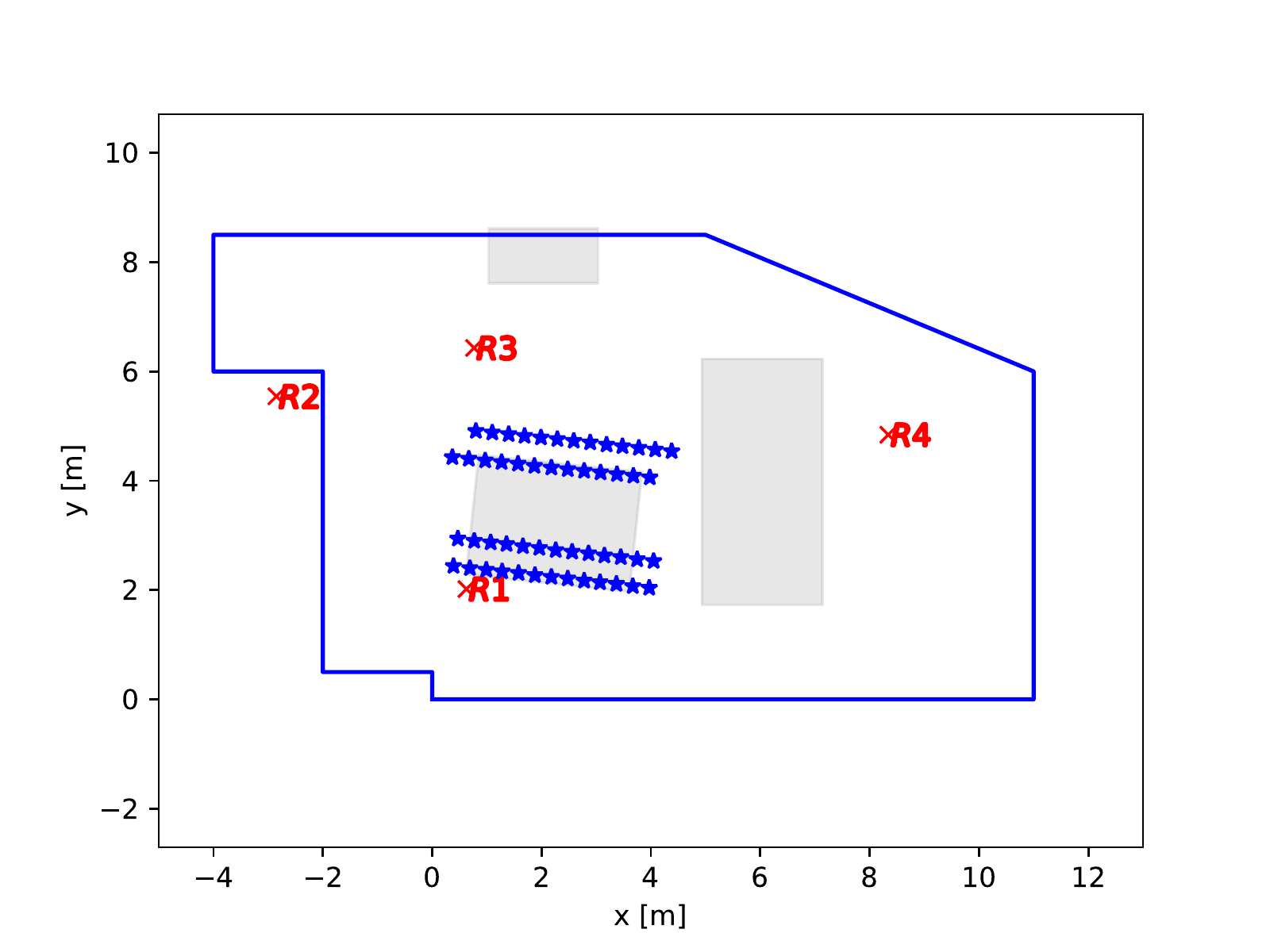}
 \caption{Plan of the  measurement room with  outer walls (solid blue),
 and tables (grey).  Locations of transmitters (blue
 star) and  receivers (red cross) are indicated. }
 \label{fig:room}
\end{figure}

\blt[tested algorithms]The proposed DNN-based classifier (DNNC) is
compared with three benchmarks.
\begin{bullets}
\blt[DBC]The first two, termed distance-based classifiers (DBCs),
decide $\ho$ iff $\| \EstVecRss-\EstVecRss'\|_q>\Thr$, where $q$ is 2
for DBC($\ell_2$) and 1 for DBC($\ell_1$). The threshold $\Thr$ is
adjusted to maximize the accuracy over the training pairs.
\blt[KMC]Along the lines of
\cite{chen2007detecting,yang2012detection}, the third benchmark is a
K-means classifier that clusters the training vectors
$\EstVecRss\NotEst{\IndEst}(\Loc_{\IndDat})$ into $\NumClu$
clusters. Then it applies the same rule as DBC($\ell_2$) over the two
feature vectors that respectively result from collecting the $\NumClu$
Euclidean distances from the given $\EstVecRss$ and $\EstVecRss'$ to
all centroids.

\blt[Performance metrics] The performance metric is the accuracy
measured on a test data set constructed as described in
Sec.~\ref{sec:data} with the $52-\NumDat$ locations that are not used
for training. The accuracy of each algorithm is averaged using Monte
Carlo simulation over the choice of the $\NumDat$ training locations
among all 52 available locations. The vector of parameters $\VecWei$
of DNNC is randomly initialized at every Monte Carlo iteration.

\end{bullets}

\end{bullets}

\cmt{description of the experiments}
\begin{bullets}
\blt[exp. 1]The first experiment investigates the
number $\NumDat$ of training locations that need to be collected to
attain a reasonable accuracy. Fig.~\ref{fig:trainingpositions}
compares the accuracy of DNNC as a function of $\NumDat$ with the
benchmarks. It is observed that (i) the accuracy of DNNC is very high
with just 45 locations, (ii) DNNC learns much more than the benchmarks
with new data. \cbr{The horizontal axis begins at $\NumDat=10$ because DNNC
needs to split the given $\NumDat$ positions into both training and
validation positions and, therefore, using $\NumDat<10$ results in
very noisy estimates of the training and validation losses;
cf.~\cite[Ch. 2]{cherkassky2007}.}

\begin{figure}[t]
 \centering
 \includegraphics[width=0.45\textwidth]{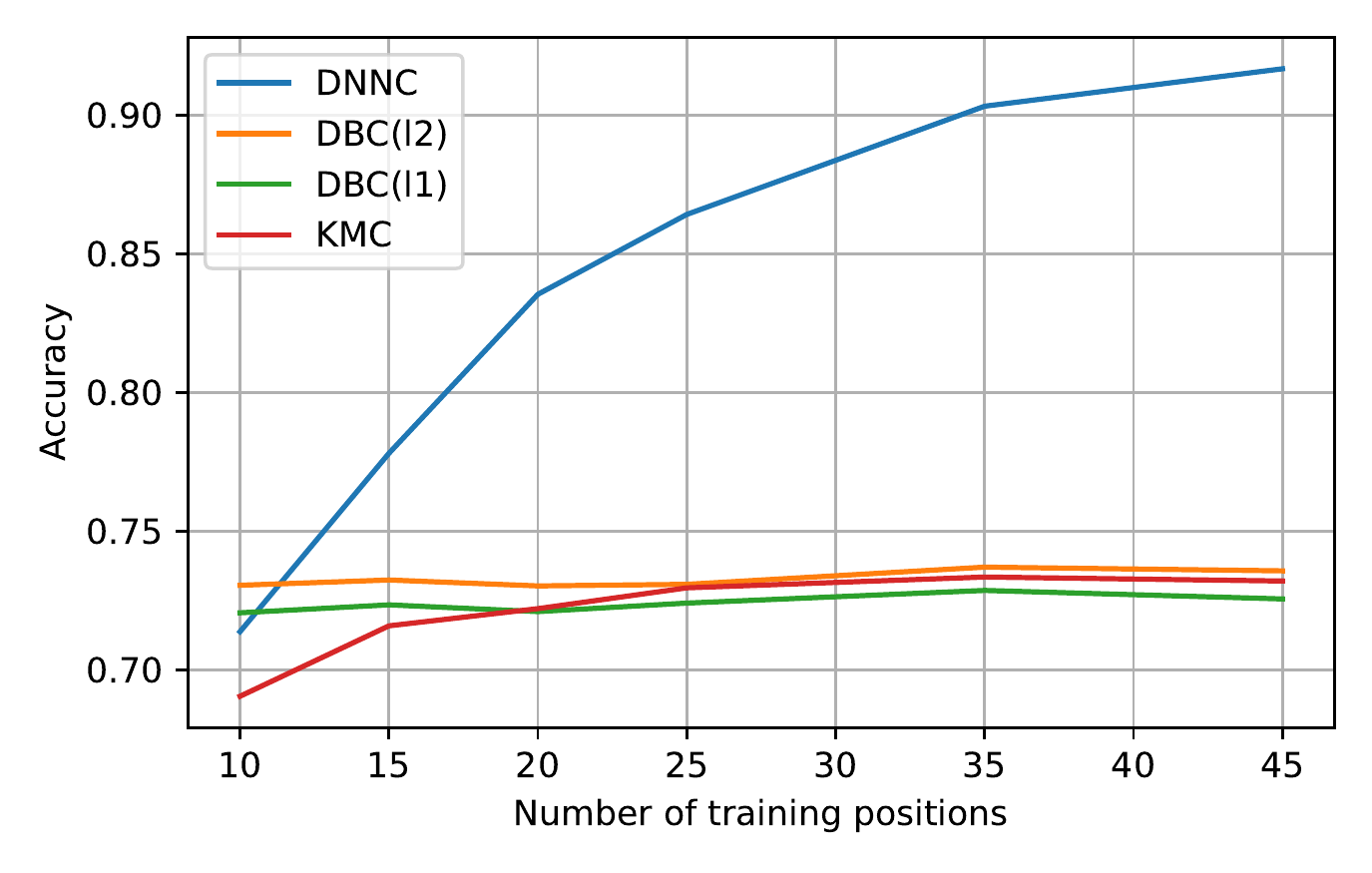}
 \caption{Accuracy vs. number of training positions $\NumDat$ for the
 proposed algorithm (DNNC) and the competing algorithms ($\NumSam=16$,
 $\NumFea=16$, 
 $\NumPai\Tra=1250$,  $\NumPai\Val=150$, $\NumDat\Tra=0.8\NumDat$,
 $\NumClu=15$).   }
 \label{fig:trainingpositions}
\end{figure}

\blt[exp. 2]The second experiment assesses the influence of the number
of features on the performance of DNNC. In this data set, the first 4
features correspond to the first receiver, the second 4 features to
the second receiver, and so on.  Fig.~\ref{fig:features} shows that
two features already result in a very high accuracy for
DNNC. Remarkably, this effect can be seen to be milder when the two
selected features are obtained by different receivers, which suggests
that there is more information in the joint distribution of the RSS
estimates obtained by nearly-located antennas than in those obtained
by distant antennas. \cbr{One may ascribe this phenomenon to the fact that
the latter are ``less synchronized'' than the former.}

\begin{figure}[t]
 \centering
 \includegraphics[width=0.45\textwidth]{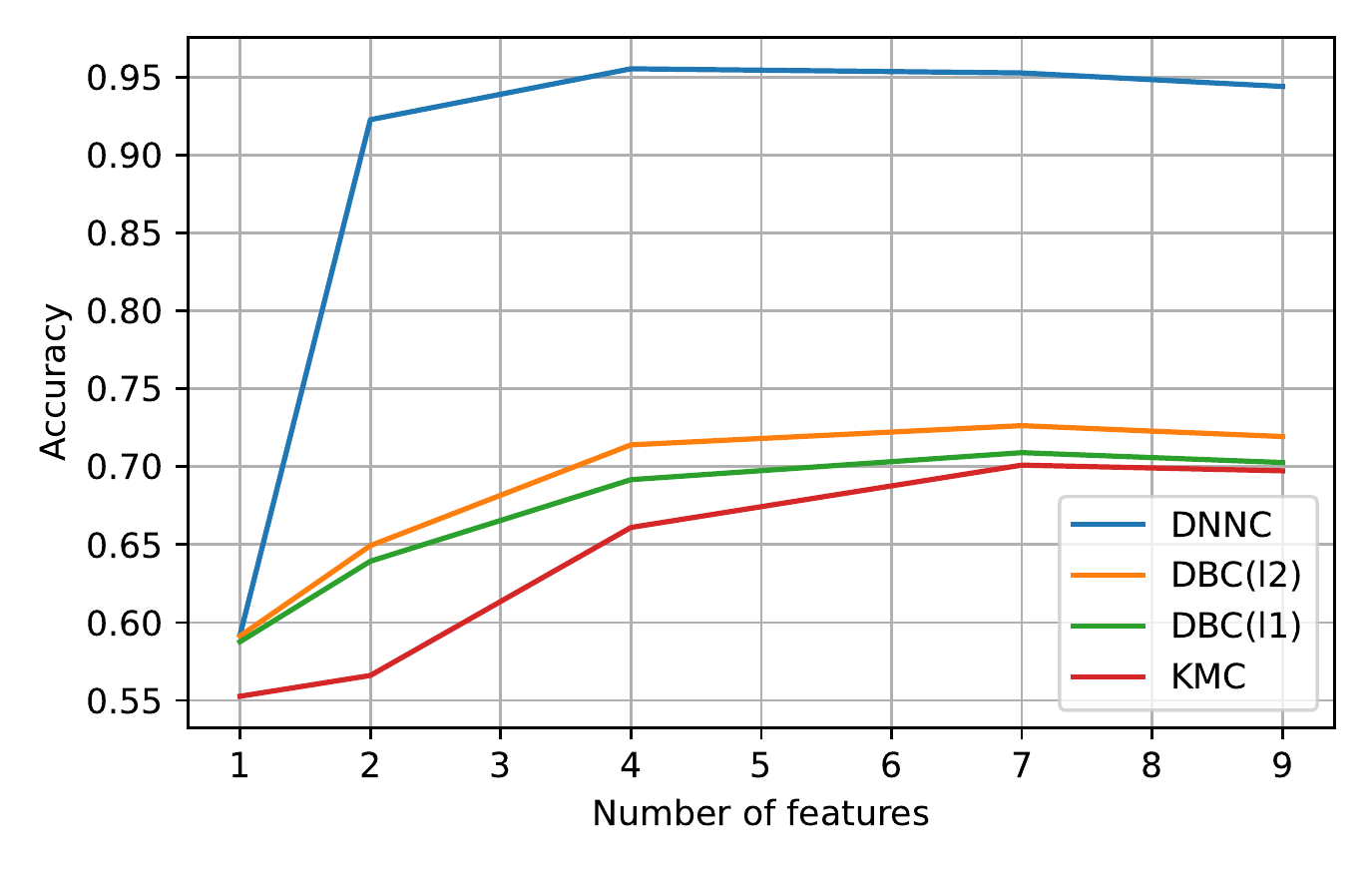}
 \caption{Accuracy vs. number of features $\NumFea$ for the
 proposed algorithm (DNNC) and the competing algorithms ($\NumSam=16$,
 $\NumPai\Tra=1250$, $\NumPai\Val=150$, $\NumDat=40$,
 $\NumDat\Tra=0.8\NumDat$, $\NumClu=15$).  }
 \label{fig:features}
\end{figure}

\blt[samples]Finally, it is worth observed that in both experiments,
the value of samples used by the RSS estimator in \eqref{eq:estrss} is
just $\NumSam=16$, which demonstrates that DNNC meets the goal of
providing a high temporal resolution. Further experiments omitted here
due to lack of space show that the accuracy of the benchmarks
decreases more abruptly than the one of DNNC if $\NumSam$ is reduced
below 16.

\end{bullets}

\section{Conclusions and Discussion}

\begin{bullets}
\blt[conclusions]This paper presented an algorithm that learns to
distinguish whether two RSS feature vectors correspond to the same
transmitter location or to a different transmitter location. The low
number of samples required by this algorithm renders it suitable to
construct a spoofing detector that can distinguish spoofing attacks
from motion of a legitimate user; cf.  Sec.~\ref{sec:intro}.  A DNN
with an architecture that ensures symmetry in the decisions was
designed and tested. Numerical experiments with real data corroborate
the high accuracy of the proposed scheme. 
Future work will target the development of a high-level algorithm that
relies on the decisions of the proposed scheme to attain robustness to
user movement. 

It is worth emphasizing that the collection of the data set entails
 small effort, since the locations of the training points
 $\Loc_\IndDat$ need not be measured. It suffices to ensure that
 $\Loc_\IndDat\neq \Loc_{\IndDat'}$ for all $\IndDat\neq\IndDat$. The
 measurements can be collected by a technician or recording signals in
 a time interval where no attacks are known (by some other means) 
 to be taking place.




\end{bullets}

\clearpage
\balance
\small
\printmybibliography
\end{document}